\documentclass[10pt,twocolumn,letterpaper]{article}

\usepackage{cvpr}              
\usepackage{multirow}
\usepackage{booktabs}
\usepackage[accsupp]{axessibility} 

\usepackage[dvipsnames]{xcolor}

\definecolor{cvprblue}{rgb}{0.21,0.49,0.74}
\usepackage[pagebackref,breaklinks,colorlinks,citecolor=cvprblue]{hyperref}

\def\paperID{16234} 
\def\confName{CVPR}
\def\confYear{2024}

\newcommand\blfootnote[1]{%
  \begin{NoHyper}%
  \renewcommand\thefootnote{}\footnote{#1}%
  \addtocounter{footnote}{-1}%
  \end{NoHyper}%
}

\title{Masked Autoencoders for Microscopy are Scalable Learners of Cellular Biology}

\author{%
  Oren Kraus$^1$
  \qquad
  Kian Kenyon-Dean$^1$
  \qquad
  Saber Saberian$^1$
  \qquad
  Maryam Fallah$^1$
  \qquad
  Peter McLean$^1$
  \and
  Jess Leung$^1$
  \qquad
  Vasudev Sharma$^1$
  \qquad
  Ayla Khan$^1$
  \qquad
  Jia Balakrishnan$^1$
  \qquad
  Safiye Celik$^1$
  \and
  Dominique Beaini$^2$
  \qquad
  Maciej Sypetkowski$^2$
  \qquad
  Chi Vicky Cheng$^1$
  \qquad
  Kristen Morse$^1$
  \and
  Maureen Makes$^1$
  \qquad
  Ben Mabey$^1$
  \qquad
  Berton Earnshaw$^{1,2}$
  \\
  $^1$Recursion\qquad
  $^2$Valence Labs
}

\begin{document}

\maketitle
\blfootnote{$^{\dagger}$An earlier version of this work appeared at the NeurIPS 2023 Generative AI and Biology Workshop \cite{kraus2023masked}.}
\blfootnote{$^{\ddagger}$Correspondence: \url{oren.kraus@recursion.com},

\url{berton.earnshaw@recursion.com}, \url{info@rxrx.ai}.}

\begin{abstract}
Featurizing microscopy images for use in biological research remains a significant challenge, especially for large-scale experiments spanning millions of images. This work explores the scaling properties of weakly supervised classifiers and self-supervised masked autoencoders (MAEs) when training with increasingly larger model backbones and microscopy datasets. 
Our results show that ViT-based MAEs outperform weakly supervised classifiers on a variety of tasks, achieving as much as a 11.5\% relative improvement when recalling known biological relationships curated from public databases. 
Additionally, we develop a new channel-agnostic MAE architecture (CA-MAE) that allows for inputting images of different numbers and orders of channels at inference time. 
We demonstrate that CA-MAEs effectively generalize by inferring and evaluating on a microscopy image dataset (JUMP-CP) generated under different experimental conditions with a different channel structure than our pretraining data (RPI-93M).  
Our findings motivate continued research into scaling self-supervised learning on microscopy data in order to create powerful foundation models of cellular biology that have the potential to catalyze advancements in drug discovery and beyond. 
Relevant code and select models released with this work can be found at: \url{https://github.com/recursionpharma/maes_microscopy}.
\end{abstract}    
\section{Introduction}
\label{sec:intro}

A fundamental challenge in biological research is quantifying cellular responses to genetic and chemical perturbations and relating them to each other~\cite{PrzybylaNatGeneticsReviews2022,VincentPhenotypic2022}. 
Image\mbox{-}based experiments have proven to be a powerful approach for exploring cellular phenotypes induced by millions of perturbations~\cite{BoutrosHCS2015}. High Content Screening (HCS) systems, which combine automated microscopy with robotic liquid handling technologies, have enabled assaying cellular responses to perturbations on a massive scale.
Recent public releases of HCS image sets, like RxRx3~\cite{fay2023rxrx3} and JUMP-CP~\cite{chandrasekaran2023jump}, consist of millions of cellular images across 100,000s of unique chemical and genetic perturbations and demonstrate the scalability of this approach. 

\begin{figure}
    \centering
    \includegraphics[width=\columnwidth]{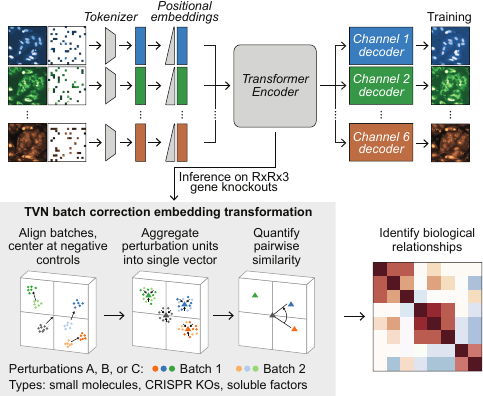}
    \caption{General depiction of the approach taken in this work. MAEs (channel-agnostic architecture depicted) learn to reconstruct HCS images, perform inference on RxRx3  \cite{fay2023rxrx3} to obtain genomic representations, and apply TVN batch correction on the embeddings to predict biological relationships.}
    \label{fig:abstract}
\end{figure}

The size of recent HCS experiments presents a unique challenge and opportunity for extracting biologically meaningful representations from these datasets. HCS images are often analyzed with customized cell segmentation, feature extraction, and downstream analysis pipelines~\cite{10.1038/nmeth.4397}.
Despite the many discoveries made using this approach~\cite{BoutrosHCS2015}, developing robust segmentation and feature extraction pipelines using proprietary or open-source software packages~\cite{10.1186/gb-2006-7-10-r100,10.1186/s12859-021-04344-9} remains challenging~\cite{ChandrasekaranHCSMLupgrade2021}. 

Alternatively, representation learning approaches do not require prior knowledge of cellular morphology and have the potential to perform significantly better on practical biological research objectives, e.g., inferring relationships between perturbations~\cite{Moshkov2022}. 
Current SOTA approaches use \textit{weakly supervised learning} (WSL)~\cite{zhou2018brief} to train models that predict the perturbations used to treat the cells in an image~\cite{caicedoWSL2018,Moshkov2022}. However, the performance of WSL models has been found to be sensitive to the strength of perturbations used~\cite{Moshkov2022}, potentially limiting the applicability of WSL to large scale datasets.

In order to overcome these limitations, we develop an alternative framework for learning representations of HCS datasets based on self\mbox{-}supervised learning (Fig.~\ref{fig:abstract}). Specifically, we train masked autoencoders (MAEs)~\cite{he2022masked} with U\mbox{-}Net and vision transformer (ViT) backbones on progressively larger HCS image sets. We show that these models, particularly MAE ViTs, are scalable learners of cellular biology, outperforming previous SOTA methods at inferring known biological relationships in whole-genome HCS screens. Specifically, we show that

\begin{itemize}[leftmargin=.5cm,itemsep=.5em]
    \item for MAEs, \textbf{recall of known biological relationships scales} with increasing model and training set sizes, while recall degrades when naively scaling WSL, 
    \item a \textbf{Fourier domain reconstruction loss} stabilizes MAE training of large ViT backbones, and
    \item employing a \textbf{novel channel-agnostic MAE ViT} helps generalize to microscopy datasets with different channel configurations.

\end{itemize}

\section{Related Work}

\label{sec:related-work}

Deep learning models have been successfully trained to perform cell segmentation \cite{Valen2016, Moen2019,Cellpose2021} and phenotype classification \cite{Kraus2016,Kraus2017,Ouyang2019,Eulenberg2017}, however these supervised learning tasks require the costly creation of segmentation masks and other labels.
Inspired by the successful use of embeddings obtained from ImageNet-trained models for other datasets and tasks \cite{Razavian2014}, researchers have used models trained on natural images to featurize HCS data with varying results \cite{TVN2017,Pawlowski2016}. 
Others \cite{Moshkov2022,sypetkowski2023rxrx1,saberian2022deemd,caicedoWSL2018} have used WSL to train convolutional networks to classify labels obtained from experimental metadata (e.g., perturbation class).
Despite obtaining SOTA results when trained on small, highly-curated image sets, we show that the performance of WSL models does not necessarily improve on larger datasets.

Vision models pretrained with \textit{self-supervised learning} (SSL) often outperform supervised models on downstream tasks \cite{he2022masked,caron2021dino,Chen2020simclr}. 
Unlike supervised pretraining~\cite{Kolesnikov2019bigtransfer}, SSL is readily applied to large datasets where labels are lacking or heavily biased.
This is useful for HCS datasets, as they contain a wide range of cellular phenotypes that are difficult for human experts to interpret and annotate. 
For example, DiNO~\cite{caron2021dino} is an SSL method that has been applied to HCS \cite{Cross-Zamirski2022dinohcs,Haslum2022metadataguided,Sivanandan2023posh,kim2023self,doron2023unbiased} data, however it relies on augmentations inspired by natural images, which may not be applicable to HCS image sets. 
Alternatively, \textit{masked autoencoders} (MAEs) \cite{he2022masked} are trained by reconstructing masked patches conditioned on unmasked patches of an image (Fig.~\ref{fig:recon}). MAEs have been successfully applied to images \cite{he2022masked}, audio \cite{Huang2022audiomae}, video \cite{Feichtenhofer2022videomae} and multimodal audio-video datasets \cite{Huang2022avmae}. However, previous attempts to train MAEs on HCS datasets have had limited success \cite{xun2023microsnoop,kim2023self}, likely due to limitations in compute resources and dataset size. 
\section{HCS Datasets} \label{datasets}

\begin{table*}[ht]
    \centering
    \begin{tabular}{lllrr}
    \toprule
    \textbf{Pretraining dataset} & Imaging modality & Perturbation type(s) & \# images & \# perturbations  \\
    \midrule
    RxRx1 \cite{sypetkowski2023rxrx1} & Cell Painting & gene KD (siRNA) &  125,510    & 1,108  \\
    RxRx1-2M                    & Cell Painting & gene KD (siRNA) &  1,650,319  & 1,108  \\
    RxRx3 \cite{fay2023rxrx3}         & Cell Painting & gene KO (CRISPR), SMC &  2,222,096  & 113,517\\ 
    RPI-52M                           & Cell Painting, Brightfield & gene KD/KO/OX, SMC, SF &  51,516,177 & 2,345,638   \\
    RPI-93M                           & Cell Painting, Brightfield & gene KD/KO/OX, SMC, SF &  92,764,542 & 3,957,400   \\
    \bottomrule
    \end{tabular}
    \caption{Summary of the HCS datasets explored for pre-training in this work. Each image in each dataset is 2,048 x 2,048 x 6 pixels. Genetic perturbations include knock-down (KD), knock-out (KO), and overexpression (OX). Non-genetic perturbations include small-molecule compounds (SMC) and soluble factors (SF; e.g. cytokines, biologics).  RPI- datasets include genetic perturbations generated with siRNA, CRISPR, and other genetic manipulation technologies.}
    \label{tab:datasets}
\end{table*}
We investigate the scaling properties \cite{zhai2022scaling} of MAE and WSL pretraining by evaluating increasingly larger models trained on five HCS microscopy datasets of different sizes, as summarized in Table~\ref{tab:datasets} (see Appendix~\ref{appendix:datasets} for additional details). In curating these datasets, we aimed to cover a broad range of biological and experimental factors that could impact a deep learning model's ability to learn transferable representations of the images. These datasets contain images captured using a six-channel proprietary implementation of the Cell Painting imaging protocol \cite{10.1038/nprot.2016.105}, which multiplexes fluorescent dyes to reveal eight broadly relevant cellular components. The RPI-52M and RPI-93M (Recursion Phenomics Imageset) datasets also include several million images obtained with Brightfield microscopy imaging. RPI-52M is a superset of RxRx1, RxRx1-2M, and RxRx3, and RPI-93M is a superset of  RPI-52M.

\section{Methods} \label{methods}
\label{sec:methods}
This section discusses the strategies we used to train deep computer vision models on our HCS image datasets (Table~\ref{tab:datasets}). During pretraining, each model receives as input 256 x 256 crops randomly sampled from images in the training set, preprocessed with channel-wise self-standardization~\cite{sypetkowski2023rxrx1}. See Appendix~\ref{appendix:modelling} for more details on training and hyperparameters.

\begin{figure*}[htbp]
    \centering
    \begin{minipage}{.25\textwidth}
        \centering
        \includegraphics[trim={0 0 0 0.8cm},clip,width=\linewidth]{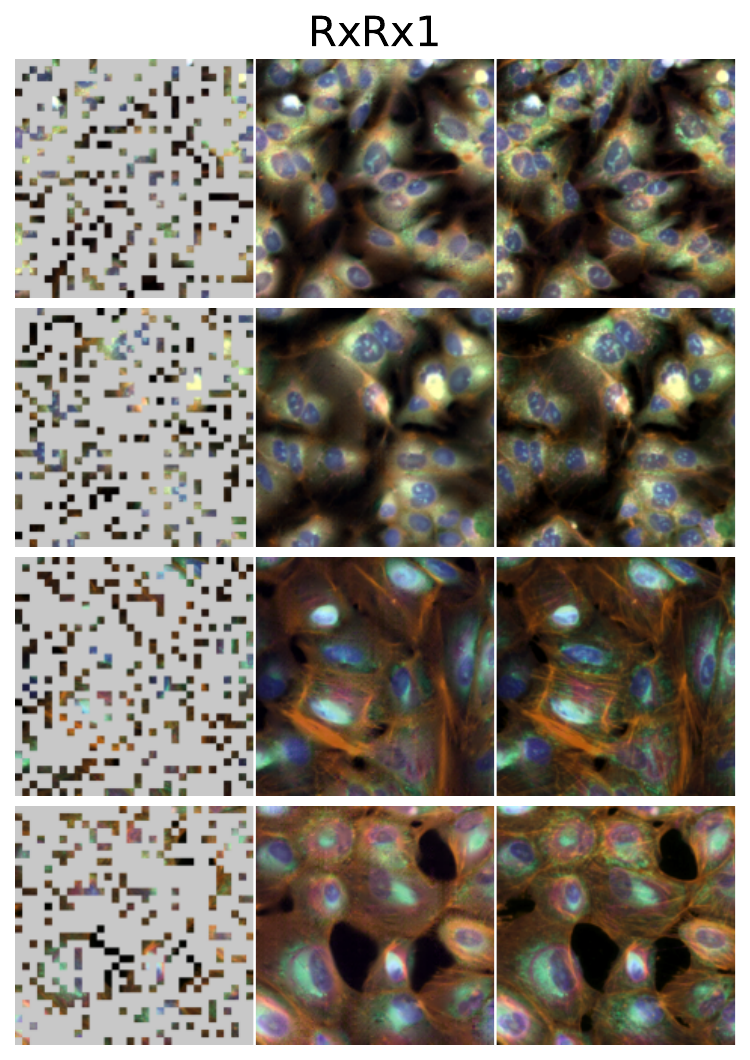}
    \end{minipage}%
    \begin{minipage}{.25\textwidth}
        \centering
        \includegraphics[trim={0 0 0 0.8cm},clip,width=\linewidth]{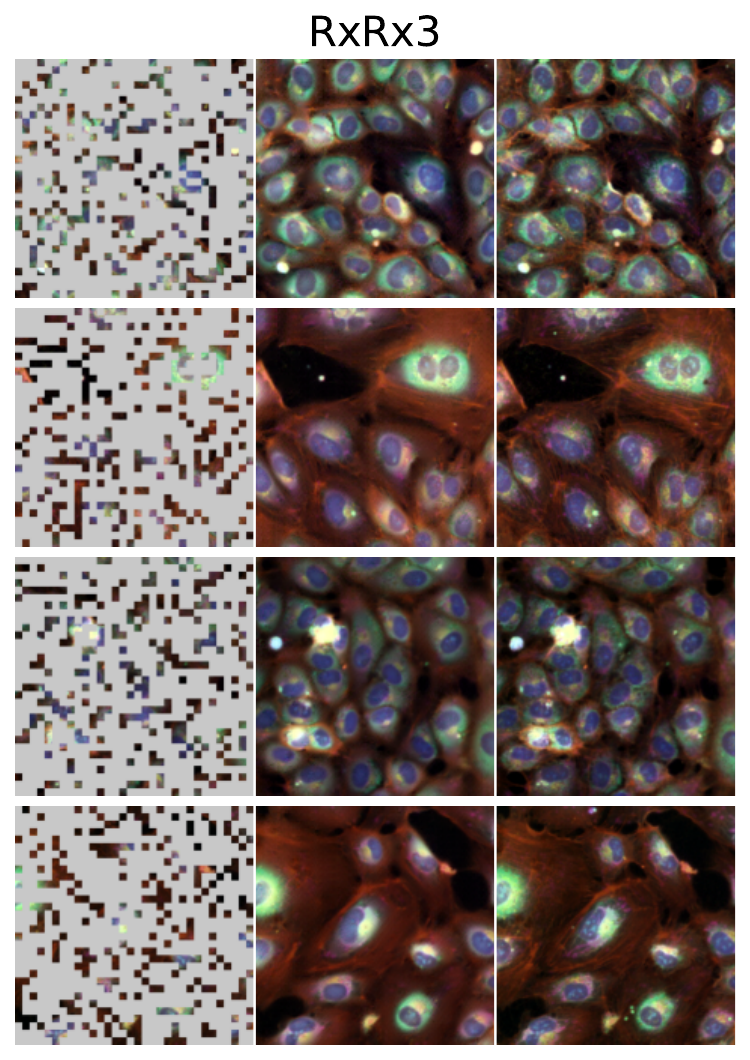}
    \end{minipage}%
    \begin{minipage}{.25\textwidth}
        \centering
        \includegraphics[trim={0 0 0 0.8cm},clip,width=\linewidth]{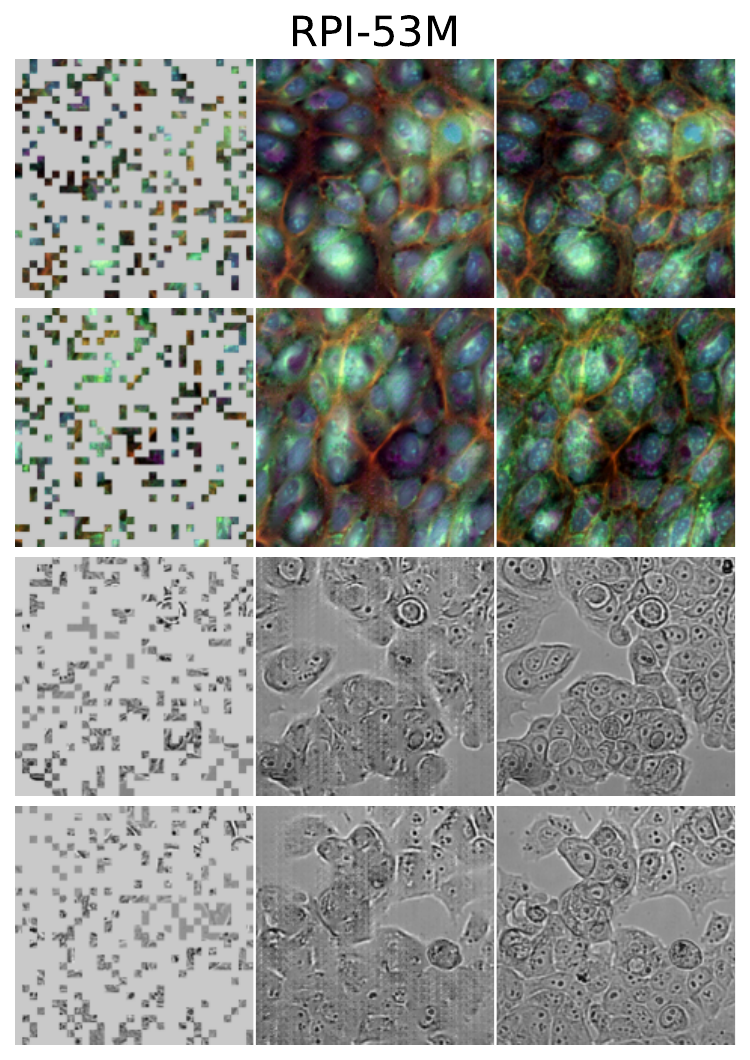}
    \end{minipage}%
    \begin{minipage}{.25\textwidth}
        \centering
        \includegraphics[trim={0 0 0 0.8cm},clip,width=\linewidth]{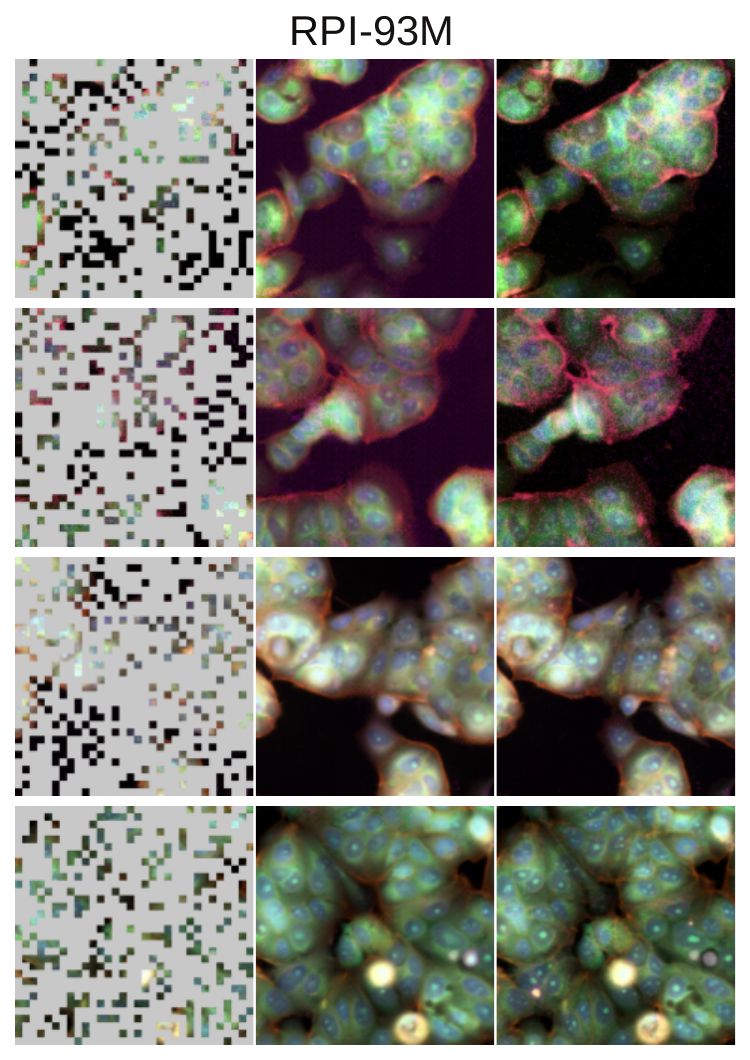}
    \end{minipage}
    \caption{Visualizing MAE ViT-L/8+ (trained on RPI-93M)  reconstructions on random \textit{validation} images from four datasets -- RxRx1, RxRx3, RPI-52M, and RPI-93M. For each dataset column, we show a triplet of the masked input (left), the reconstruction (middle), and the original (right); for this model, we randomly mask 75\% of the 1,024 8x8 patches constructed from the 256 x 256 center crop of the full image. Images are taken from wells on the same experimental plate, rows alternate between randomly sampled negative control and perturbation conditions (see Fig.~\ref{fig:abstract}).}
    \label{fig:recon}
\end{figure*}

\subsection{Weakly supervised learning}\label{wsl}

We train WSL models to classify \textit{perturbations}; i.e., to predict the genetic or chemical perturbation applied to the cells (e.g., siRNA knockdown, CRISPR knockout, or small molecule) in a random crop of a training image as input. 

We reimplement the 28-million parameter \mbox{DenseNet-161} backbone proposed in \citet{sypetkowski2023rxrx1}, trained to predict cellular perturbations and producing 128-dimensional embeddings from a two-layer MLP neck before the classification logits. We also trained model variants that produce 1,024-dimensional embeddings. We trained such models with and without adaptive batch normalization (AdaBN), an architectural technique to enable domain adaptation \cite{li2018adaptive}.
Our AdaBN-based DenseNet-161 classifiers are implemented with Ghost BatchNorm \cite{hoffer2017train} in order to train with larger batch sizes.

We also trained WSL models with vision transformers (ViT-B/16 and ViT-L/16) \cite{dosovitskiy2020image}, described further in the following sections. Our ViT classifiers use the embedding of the class token from the final layer as the representation of the image crop (we observed minimal difference in downstream performance between using the class token embedding versus averaging over patch embeddings).

\subsection{Masked autoencoders}
We train and evaluate MAEs with convolutional and transformer backbones of different sizes, depending on the scale of the training set. We provide example reconstructions on our pretraining validation sets in Figure~\ref{fig:recon}, and additional reconstructions in the Appendix~\ref{appendix:recons}. 

We adapt U-Nets~\cite{ronneberger2015u} for use as masked autoencoders (MU-Nets) by training to reconstruct masked sections of input images. We train MU-Nets as described in \citet{xun2023microsnoop} and report results for MU-Net-M and MU-Net-L, which have 52- and 135-million parameters, respectively. MU-Net-M's downsampling schedule is 32/64/128/256/512, while MU-Net-L incorporates an additional block of size 1,024. In each case, the decoder mirrors the encoder.

We train vision transformers~\cite{dosovitskiy2020image,steiner2021train,dehghani2023scaling,zhai2022scaling} as MAEs following the implementation in \citet{he2022masked}. 
We report results for ViT-S, ViT-B, and ViT-L encoders \cite{dosovitskiy2020image}, containing 22-, 86-, and 304-million parameters, respectively, and producing 384-, 768-, and 1,024-dimensional embeddings respectively. We explore the use of 8x8 and 16x16 patch sizes and 75\% and 25\% mask ratios (Fig.~\ref{fig:recon}), respectively. A 25-million parameter decoder \cite{he2022masked} is used for patch reconstructions. Note that 8x8 patches induce a sequence length 4 times greater than 16x16 patches and are thus more computationally expensive. Our MAE ViTs use the average of patch embeddings from the final layer of the encoder as the embedding of the image crop.

We observed (Fig. \ref{fig:fourier}) an interesting behavior when training large MAE-ViTs on our largest datasets. Early in training, after a steep initial descent in loss, the model encountered an apparent saddle point region in the parameter landscape. When trained long enough, we could surpass that region and ``double-dip'' the loss curve after many million crops are seen (depending on model and dataset size). We found that training dynamics and downstream performance benefited from large batch sizes of up to 16,384 image crops and using the Lion optimizer \cite{chen2023symbolic}, versus the typical choices of batch size and AdamW optimizer~\cite{SSLcookbook2023}.

\subsubsection{Fourier domain reconstruction loss} \label{sec:fourier}
\begin{figure}[t]
    \centering
    \includegraphics[width=\columnwidth]{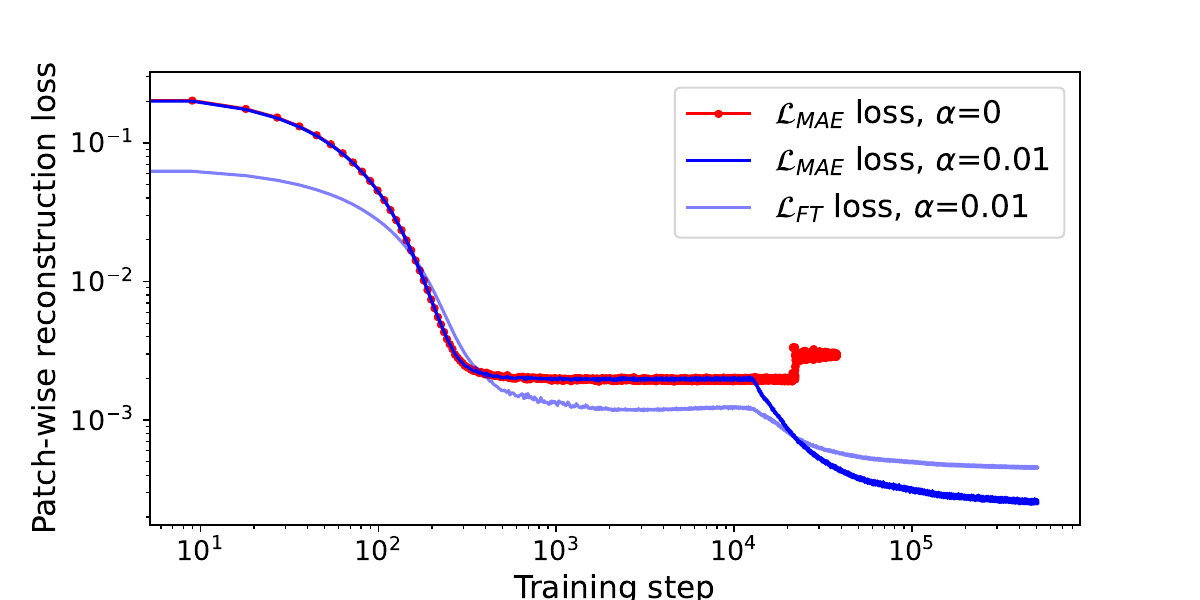}
    \caption{Example reconstruction loss curves (log-log scale) training a CA-MAE ViT-L/16, with and without Fourier domain reconstruction loss (same random seed), on RPI-93M; similar results hold for other large MAE ViTs across multiple runs. Training with $\mathcal{L}_{F}$ at $\alpha=0.01$ (Eq.~\ref{eq:combined}) enables surpassing the saddle-point region.}
    \label{fig:fourier}
\end{figure}

Even with the training strategies described above, our largest models with many tokens, such as ViT-L/8, diverged early during training. We also observed that reconstructions lacked the kind of texture prediction that characterize microscopy images, consistent with the original MAE results in which high-frequency textures were not reconstructed well~\cite{he2022masked}. We therefore added an additional reconstruction loss in the Fourier domain~\cite{xie2022masked} to encourage the model to better reconstruct the textures of cellular morphology, which also facilitated more reliable navigation of the loss landscape for reconstruction in general. 

MAEs are trained with mean squared error ($L_2$) reconstruction loss at the patch level only on the masked patches. Formally, given $P$ masked patches for an individual sample, the patch's image pixels $y_p$ and the model's reconstruction of the patch $y'_p$:
\begin{equation}\label{eq:mae}
\mathcal{L}_{MAE} = \frac{1}{P} \sum_{p=1}^{P} L_2(y_p, y'_p).
\end{equation} 
We incorporated an additional loss term based on the fast Fourier transformation, $\mathcal{F}$, following the standard reconstruction loss in Eq.~\ref{eq:mae}, calculated on masked patches only:
\begin{equation}\label{eq:fft}
\mathcal{L}_{FT} = \frac{1}{P} \sum_{p=1}^{P} L_1( |\mathcal{F}(y_p)|, |\mathcal{F}(y'_p)|).
\end{equation} 
This loss term incentivizes the model to minimize the mean absolute error ($L_1$) between the original and reconstructed patches in the frequency domain.

Finally, we combine Eqs.~{\ref{eq:mae}} and~\ref{eq:fft} as follows:
\begin{equation}\label{eq:combined}
    \mathcal{L}_{MAE+} = (1 - \alpha) \mathcal{L}_{MAE} + \alpha \mathcal{L}_{FT},
\end{equation}
where the hyperparameter $\alpha \in (0, 1)$. All models indicated with a + (e.g., ViT-L/8+) are trained using this loss function. We found that setting $\alpha=0.01$ worked effectively. As illustrated in Figure~\ref{fig:fourier}, we found that training with this loss term consistently resulted in a stable double-descent in loss.

\begin{figure*}[ht]
    \centering
    \includegraphics[width=\textwidth]{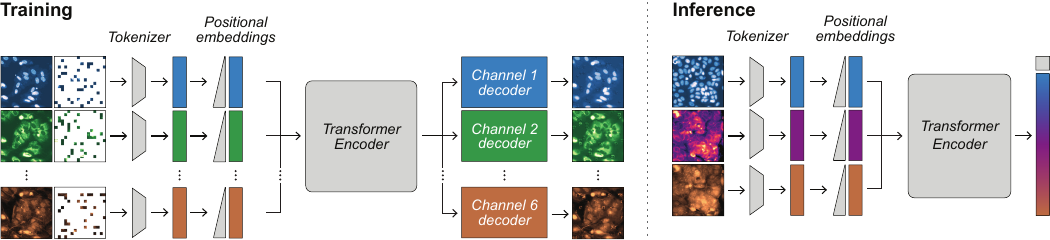}
    \caption{Channel-agnostic MAE (CA-MAE). This architecture enables transferring ViT encoders trained using MAEs from one set of channels to another. \textit{Left}: CA-MAE training (ViT-L/16+, 85\% mask) in which an input tensor is split into individual channels and a shared linear projection (Tokenizer) is applied to each channel, followed by the addition of positional embeddings per channel. \textit{Right}: the trained ViT encoder can then be used to embed images with different sets, ordering, and/or numbers of channels (3 shown here) by using the class token, averaging all the patch embeddings, or averaging the patch embeddings from each channel separately and concatenating them.}
    \label{fig:CA-MAE}
\end{figure*}

\subsubsection{Channel-agnostic MAEs} \label{camae}
Microscopy images captured by HCS can vary significantly across experiments and labs, often containing different numbers of channels and different cellular objects stained in each channel. Although many labs have aligned on the Cell Painting protocol~\cite{10.1038/nprot.2016.105}, there are still variations between experimental implementations, with some protocols having 5 or 6 of the fluorescent morphology stains, and others adding brightfield or experiment-specific channels. Standard convolutional-~\cite{alexnet} or vision transformer-based~\cite{dosovitskiy2020image} architectures require input images to have a consistent set of channels between training and test settings. 

In an effort to develop an architecture that can transfer to a different number and set of channels at test time, we developed the \textit{channel-agnostic ViT} architecture (\mbox{CA-MAE}). This architecture was inspired by recent work on multimodal MAEs~\cite{bachmann2022multimae,geng2022multimodal}, specifically~\citet{bachmann2022multimae}, in which RGB images, scene depth and semantic segmentation are considered separate modalities that train a single ViT-based MAE. Our implementation treats each channel as a separate modality, creating \(C \times N\) tokens where $C$ is the number of channels and $N$ is the number of patches defined by $N = HW/P^2$, where $(H, W)$ is the resolution of the original image, and $(P, P)$ is the resolution of each image patch. To make the model agnostic to the number and set of channels at test time, we apply a single shared linear projection and the same positional embeddings to all channels based on the standard sine-cosine functions~\cite{he2022masked}.
We apply the masking ratio to the resulting \(C \times N\) tokens, producing different masks for each channel. During training, we use separate decoders for each channel similar to the separate decoders used for each modality in~\citet{bachmann2022multimae}. We use a 75\% (ViT-B/16) or 85\% (ViT-L/16) masking ratio. Figure~\ref{fig:CA-MAE} describes this architecture in detail. 
\section{Results}
\label{sec:results}

We evaluated our models based on their ability to identify biological relationships as well as predict aggregated single cell features~\cite{10.1186/s12859-021-04344-9}. 

\begin{table}[t]
\centering
\caption{Impact of batch correction methods for RPI-93M MAE ViT-L/8+; findings are similar for other models. Recall of known relationships in top and bottom 5\% of cosine similarities on CORUM/hu.MAP/Reactome/StringDB databases.}
\begin{tabular}{lc}
\toprule
\textbf{Transformation method} & \textbf{Recalls} \\
\midrule

No transformation & .124/.124/.096/.135\\
PCA & .126/.122/.102/.134\\
Center by plate & .449/.361/.184/.350\\
Center by experiment & .455/.365/.186/.353\\
Standardize by plate & .456/.367/.187/.359\\
Standardize by experiment & .460/.370/.188/.359\\
PCA+Standardize by plate & .614/.435/.261/.477\\
PCA+Standardize by experiment & .614/.435/.258/.477\\

TVN & \textbf{.622}/\textbf{.443}/\textbf{.267}/\textbf{.484} \\
\bottomrule
\end{tabular}
\label{table:tvn}
\end{table}

\subsection{Predicting biological relationships}\label{sub_section:BMDB}

\begin{table*}[t]
\centering
\caption{Recall of known relationships in top and bottom 5\% of cosine similarities by model, pretraining set, and database. All results are computed on RxRx3 after applying TVN and chromosome arm bias correction. Results include simple baselines, intermediate model checkpoints, ablations, and performant WSL/SSL models. MAEs with + are trained with Fourier domain reconstruction loss, $\alpha=0.01$ (Eq.~\ref{eq:combined}).} 
\begin{tabular}{llcccc}
\toprule
\textbf{Model backbone} & \textbf{Pretraining dataset} & CORUM & hu.MAP & Reactome & StringDB \\
\midrule
\multicolumn{1}{l}{\textit{Simple baselines}} \\
Random 1024-dim embeddings  & N/A & .100 & .100 & .100 & .100 \\
Pixel intensity statistics & N/A & .280 & .260 & .160 & .270\\  
\midrule

\multicolumn{1}{l}{\textit{ImageNet-pretrained classifiers}} \\
ViT-S/16 &  Imagenet-21k \cite{ridnik2021imagenet} &  .494 & .348 & .213 & .388 \\
ViT-B/16 &  Imagenet-21k \cite{ridnik2021imagenet} & .511 & .344 & .216 & .395 \\
ViT-B/8  &  Imagenet-21k \cite{ridnik2021imagenet} & .472 & .324 & .203 & .369 \\
ViT-L/16 &  Imagenet-21k \cite{ridnik2021imagenet} & .531 & .360 & .228 & .409 \\
\midrule

\multicolumn{1}{l}{\textit{Weakly supervised models}} \\
DenseNet-161 & RxRx1 \cite{sypetkowski2023rxrx1} & .383 & .307 & .190 & .330\\
DenseNet-161 w/ AdaBN    & RxRx1 \cite{sypetkowski2023rxrx1} & .485 & .349 & .228 & .417\\
DenseNet-161 w/ AdaBN   & RxRx3 \cite{fay2023rxrx3}         & .461 & .303 & .188 & .377\\
DenseNet-161 w/ AdaBN (1024-dim)  & RxRx1 \cite{sypetkowski2023rxrx1} & .502 & .363 & .220 & .422\\
DenseNet-161 w/ AdaBN (1024-dim)  & RxRx3 \cite{fay2023rxrx3}         & .520 & .350 & .207 & .413\\
ViT-B/16 & RxRx1 \cite{sypetkowski2023rxrx1} & .505 & .348 & .218 & .408 \\  
ViT-L/16 & RxRx3 \cite{fay2023rxrx3}  & .532&	.353&	.196&	.402 \\
ViT-L/16 & RxRx1-2M & .568 & .397 & .255 & .472\\

\midrule

\multicolumn{1}{l}{\textit{MU-Nets}} \\
MU-Net-L      & RxRx3 \cite{fay2023rxrx3}     & .566 & .374 & .232 & .427 \\
MU-Net-L      & RPI-52M    & .576 & .385 & .238 & .443 \\
MU-Net-L      & RPI-93M    & .581 & .386 & .247 & .440 \\
\midrule

\multicolumn{1}{l}{\textit{Intermediate MAE ViT checkpoints}} \\
MAE ViT-L/8+ (epoch 1) & RPI-52M & .524 & .357 & .216 & .405 \\  
MAE ViT-L/8+ (epoch 25) & RPI-52M & .595 & .411 & .254 & .461\\ 
MAE ViT-L/8+ (epoch 46) & RPI-52M & .605 & .424 & \textbf{.267} & .474\\ 
\midrule

\multicolumn{1}{l}{\textit{MAE ViTs}} \\
MAE ViT-B/16  & RxRx3 \cite{fay2023rxrx3}      & .565 & .387 & .232 & .435 \\
MAE ViT-B/16  & RPI-52M    & .540 & .373 & .234 & .416 \\
MAE ViT-B/8   & RPI-52M    & .601 & .404 & .251 & .459 \\
MAE ViT-L/16  & RxRx3 \cite{fay2023rxrx3}     & .560 & .374 & .231 & .427 \\
MAE ViT-L/16  & RPI-52M    & .607 & .414 & .258 & .460 \\
MAE ViT-L/16+ & RPI-52M    & \textbf{.626} & .425 & .260 & .468 \\
MAE ViT-L/8+  & RPI-93M    & .622 & \textbf{.443} & \textbf{.267} & \textbf{.484} \\ 

\midrule

\multicolumn{1}{l}{\textit{Channel-agnostic MAE ViTs}} \\
CA-MAE ViT-B/16  & RPI-52M & .587 & .404 & .257	& .459 \\
CA-MAE ViT-B/16+ & RPI-52M & .586 & .398 & .249 & .455 \\
CA-MAE ViT-L/16+ & RPI-93M & .614 & .424 & .264 & .478 \\

\bottomrule
\end{tabular}
\label{table:baselines}
\end{table*}

\begin{figure*}[ht]
    \centering
    \begin{minipage}{0.49\textwidth}
        \centering
        \includegraphics[width=0.99\textwidth]{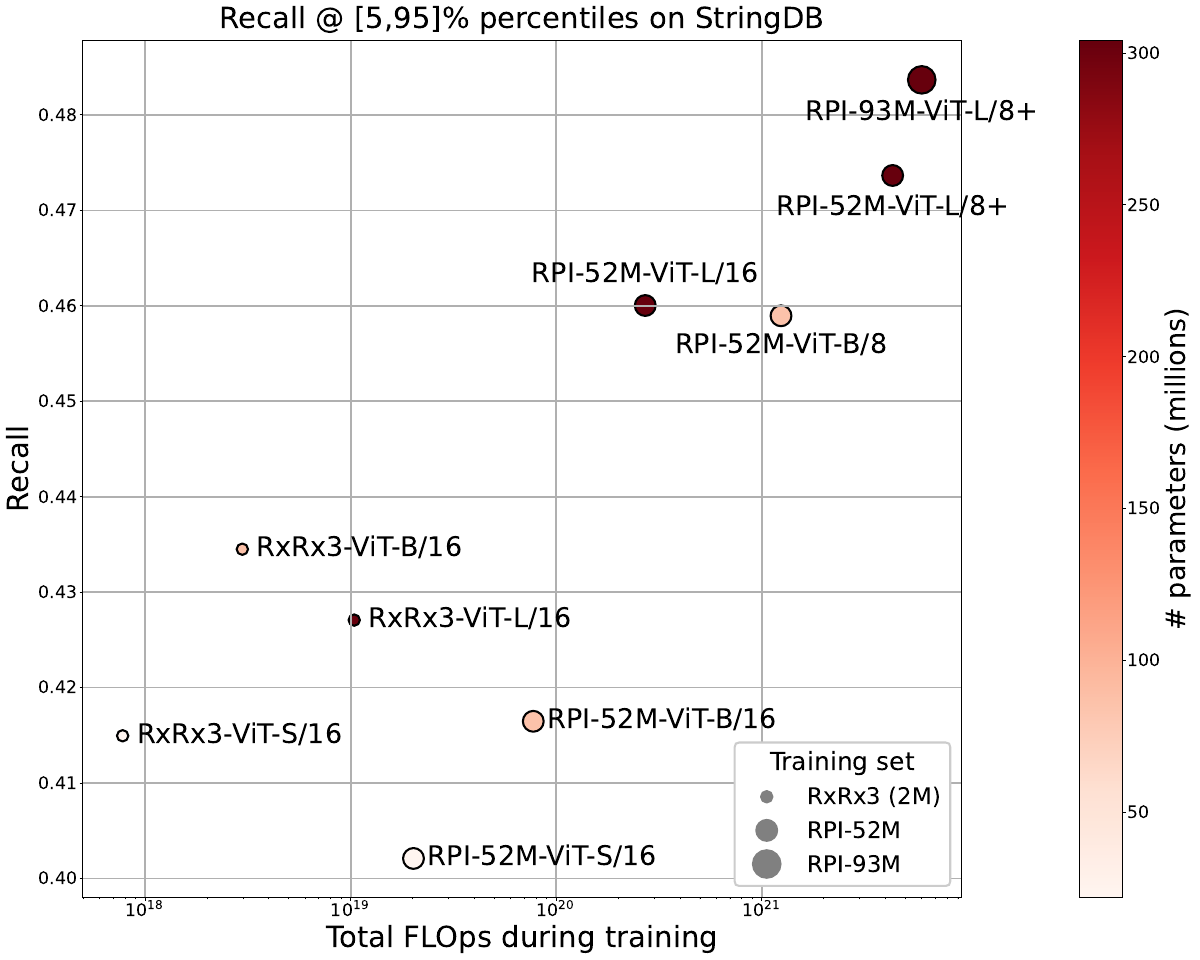}  
    \end{minipage}%
    \begin{minipage}{0.02\textwidth}
        \ 
    \end{minipage}%
    \begin{minipage}{0.49\textwidth}
        \centering
        \includegraphics[width=0.99\textwidth]{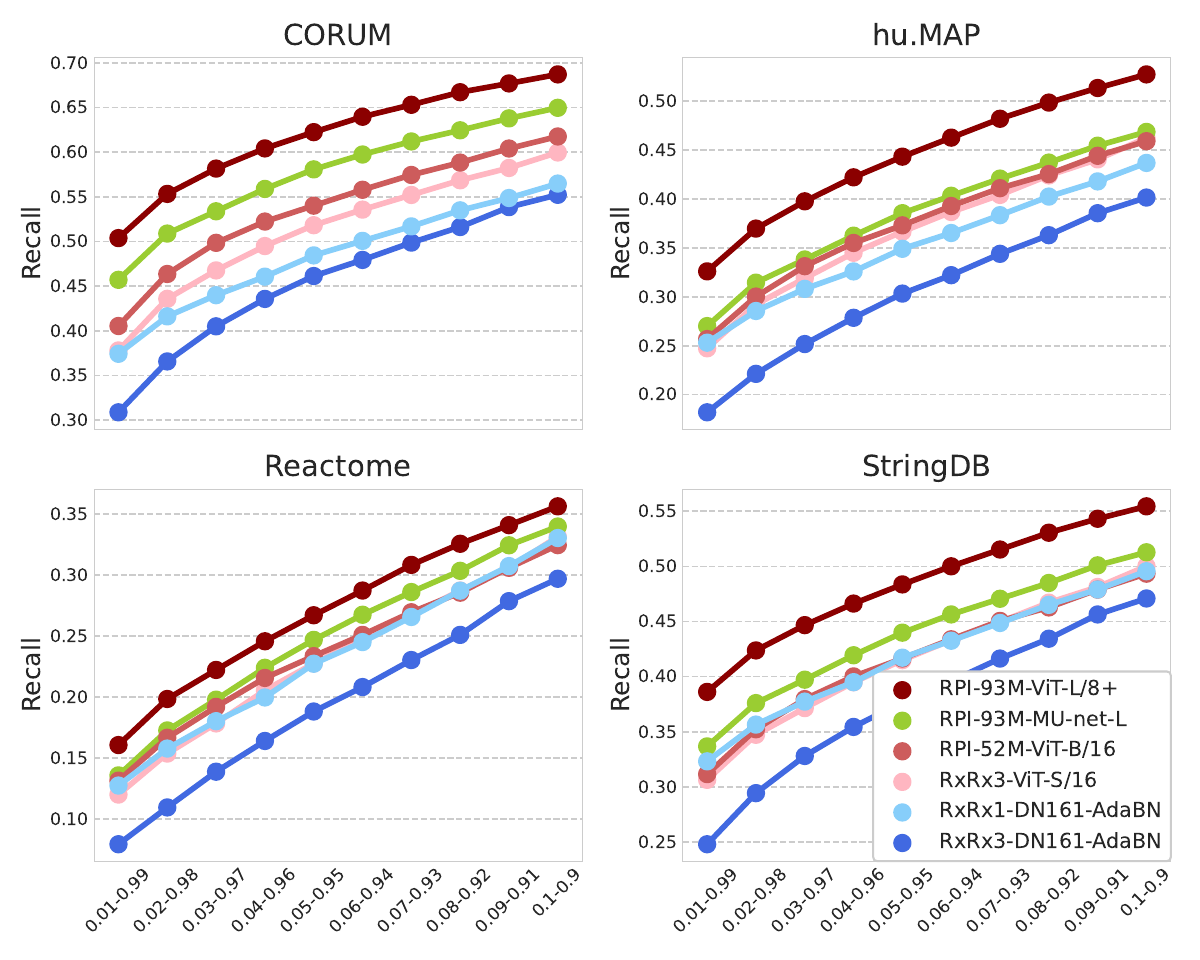} 
    \end{minipage}
    \caption{Results for select MAE ViTs taken from Table~\ref{table:baselines}. \textit{Left}: StringDB recall as a function of number of training FLOps. \textit{Right}: Recall across different cosine similarity percentiles on each database. Similar results hold for other models on other datasets. 
    }
    \label{fig:scaling}
\end{figure*}

A valuable use of large-scale HCS experiments is to perform large-scale inference of biological relationships between various types of perturbations.
We evaluate each model's ability to recall known relationships by using the multivariate metrics described in \citet{celik2022biological}. We correct for batch effects using \textit{Typical Variation Normalization} (TVN) \cite{TVN2017,celik2022biological}, and also correct for possible chromosome arm biases known to exist in CRISPR-Cas9 HCS data \cite{lazar2023high}. Table~\ref{table:tvn} shows the impact of other batch correction techniques on relationship prediction.

To predict biological relationships, we compute the aggregate embedding of each perturbation by taking the spherical mean over its replicate embeddings.
We use the cosine similarity of a pair of perturbation representations as the relationship metric, setting the origin of the space to the mean of negative controls.
We compare these similarities with the relationships found in the following public databases: CORUM~\cite{CORUM}, hu.MAP~\cite{hu_map}, Reactome~\cite{REACTOME}, and StringDB~\cite{stringdb} (with $>$95\% combined score).

Table~\ref{table:baselines} reports the recall of known relationships amongst the top and bottom 5\% of all cosine similarities between CRISPR knockout representations in RxRx3 \cite{fay2023rxrx3}. This required embedding approximately 140 million image crops and aggregating them by gene. As expected, random baselines recall $\sim$10\% of known relationships in each database (since recall is calculated from 10\% of all cosine similarities). A baseline using 30 different pixel intensity statistics as image features already recalls relationships surprisingly well compared to random. Just as surprising, pretrained ImageNet models outperform most WSL models trained on HCS datasets.
The one exception is ViT-L/16 trained on RxRx1-2M. RxRx1-2M is a dataset carefully curated to contain a large number of distinct perturbations with strong, consistent phenotypes across many cell types. The relative improvement this model achieves over training on RxRx3 suggests that implementing WSL on HCS data requires the training dataset to be curated for high-quality classes. However, this is resource intensive, experimentally and computationally, and would need to be repeated for every new HCS assay.

As previously described, we train MU-Nets and MAE ViTs of various sizes on increasingly larger datasets. Table \ref{table:baselines} shows that MAEs outperform the pretrained ImageNet and WSL models, especially when we scale up to larger model and training set sizes. For example, our best MAE model, ViT-L/8+ trained on RPI-93M, achieves a 11.5\% relative improvement over the best WSL model, ViT-L/16 trained on RxRx1-2M, when recalling known biological relationships in hu.MAP. For reasons mentioned in the previous paragraph, we did not train WSL models on datasets larger than RxRx3.
We also show the performance of intermediate MAE ViT checkpoints and observe that, as training progresses, both the reconstruction of validation images (training loss for epochs 1, 25, and 46 was 2.4e-3, 4.4e-4, and 4.1e-4, respectively) and recall of known biological relationships improve. This indicates that image reconstruction is an appropriate proxy task for  capturing biological information for use in downstream tasks of interest. 

\textbf{CA-MAE.} 
Table \ref{table:baselines} shows results for three channel-agnostic MAEs (Sec.~\ref{camae}). Note that CA-MAE ViT-B/16 significantly outperforms the MAE ViT-B/16 when trained on RPI-52M, suggesting that these architectures can offer improved performance over standard MAE ViTs. Moreover, CA-MAEs enable generalizing to datasets with different numbers of channels (see Sec. \ref{subsec:transfer_jumpcp}). We did not scale CA-MAE to the best performing MAE ViT-L/8+ architecture due to the large number of tokens generated by this architecture (6,144 for 6-channel images). We leave exploring techniques to address large token sequences in training MAEs (e.g., SWIN \cite{liu2021swin,liu2022swin,zhang2022survey} or dilated attention \cite{hassani2022dilated}) to future work.

\subsection{MAEs are scalable learners of cellular biology} \label{subsection:scalable}

In Figure~\ref{fig:scaling} we see that recall strongly correlates with the number of training FLOps, a function of both model and training set size (see Appendix \ref{sec:additional} for similar trends on other databases).
We also see that the relative performance of different pretrained models on this metric is preserved for different choices of similarity percentiles.
Our overall best model, RPI-93M MAE ViT-L/8+, is an MAE ViT-L using 8 x 8 patching, 75\% mask ratio, and trained with the Fourier domain reconstruction loss (Eq.~\ref{eq:combined}) on 128 A100 GPUs for over 20,000 GPU hours on the largest dataset, RPI-93M.

\subsection{Transfer to JUMP-CP}
\label{subsec:transfer_jumpcp}
\begin{table}
\centering
\caption{Perturbation detection and siblings 
retrieval on the JUMP-CP dataset, measured in fraction retrieved. Values are averaged ($\pm$~standard deviation) over cell types, modalities, and time-points.
}
\begin{tabular}{lcc}
\toprule
\textbf{Model backbone, dataset} & \textbf{Pert.} & \textbf{Siblings}\\
\midrule
CellProfiler \cite{10.1186/s12859-021-04344-9} & .53 \small$\pm $.30 & \textbf{.13} \small$\pm$.07  \\
ViT-L/16, ImageNet-21k \cite{ridnik2021imagenet}   & .88 \small$\pm$.09       & .06 \small$\pm$.03  \\ 
WSL ViT-L/16, RxRx1-2M  & .84  \small$\pm$.08        & .02 \small$\pm$.02 \\
MAE ViT-L/8+, RPI-93M     & .78   \small$\pm$.13       & .03 \small$\pm$.03  \\
CA-MAE ViT-L/16+, RPI-93M & \textbf{.95} \small$\pm$.05& .02 \small$\pm$.02  \\
\bottomrule
\end{tabular}
\label{table:jumpcp-benchmark}
\end{table}

To further evaluate the transferability of our models, we inferenced CPJUMP1, a subset of the JUMP-CP~\cite{chandrasekaran2023jump} dataset, and ran the corresponding benchmarking tasks introduced in~\citet{chandrasekaran2022three}. This dataset includes Cell Painting and Brightfield images of two different cell types with $\sim$130K unique perturbations and consists of two primary tasks, perturbation retrieval and sibling retrieval, where siblings represent similar but distinct perturbations. For both tasks, cosine similarity between samples is measured for individual perturbations or siblings, and Average Precision ($AP$) is measured against a null of negative control samples. Permutation testing is used to establish the significance of the $AP$ values, which are then false discovery rate-adjusted to yield q values with a cut-off of 5\% for being considered as retrieved.

Some adaptations for image embedding and data normalization were necessary compared to \citet{chandrasekaran2022three}, including our use of TVN on the negative controls to normalize the embeddings rather than \textit{robustize MAD}. Additionally, use of the WSL ViT-L/16 and MAE ViT-L/8+ models required mapping the JUMP-CP stains to those of the training set and duplicating one channel to match the model’s expected six. Meanwhile, the CA-MAE model jointly embedded the five Cell Painting channels and three Brightfield channels, despite being only trained on unpaired six-channel inputs.

We observe significantly improved performance of deep learning models on the perturbation retrieval task compared to CellProfiler \cite{10.1186/s12859-021-04344-9}, while having smaller variability across cell types, modalities, and time-points, indicating that normalized embeddings from these models consistently represent perturbations despite plate and well variations (Table~\ref{table:jumpcp-benchmark}).

In contrast, we note the lower performance of the normalized MAE model embeddings on the sibling retrieval task, where experimentally related pairs of perturbations are less similar compared to CellProfiler features. These observations are consistent with the hypothesis that MAE-trained models produce highly-resolved representations of cellular images that, in this case, are also capable of differentiating even biologically or chemically related perturbations. This illustrates the need to further develop fine-tuning strategies, or alignment methods techniques to increase performance on application-specific tasks, such as relatability among similar reagents in spite of phenotypic variation (as seen here), or other biologically-relevant research objectives like identifying genetic interactors or compound mechanisms of action. 

\subsection{Comparison with external platforms}

We compare these models with recent results from an alternative HCS platform combining pooled CRISPR screening with Cell Painting \cite{Sivanandan2023posh}. Table~\ref{table:poshcp-comparison} reports recall at 5\% FPR in StringDB on three gene sets defined in \citet{Sivanandan2023posh}. The ViT-L/8+ MAE trained on RPI-93M yields a minimum 20\% relative improvement in gene set performance over CP-DiNO 1640 (ViT-S/8), which was trained on $\sim$1.5M single-cell images. We note the significant differences in assay technology, cell lines, and modeling methodology between the two platforms, making their direct comparison impossible using this metric. Nonetheless, we hope this comparison brings the field closer to an accepted set of benchmarks for evaluating models trained on HCS datasets.

\begin{table}[t]
\centering
\caption{Recall (at 5\% false positive rate) of StringDB relationships for select models on three different gene sets PoC-124/MoA-300/DG-1640 as defined in \citet{Sivanandan2023posh}.}
\begin{tabular}{llccc}
\toprule
\textbf{Model backbone}& \textbf{Training data} & \textbf{Recalls}\\
\midrule
WSL DN161 w/ AdaBN & RxRx1 \cite{sypetkowski2023rxrx1} & .79/.\textbf{24}/.15 \\
MAE ViT-S/16 & RxRx3 \cite{fay2023rxrx3} & .74/.19/.14 \\
MU-net-L & RPI-52M  & .79/.20/.15 \\
MAE ViT-L/8+ & RPI-93M & .\textbf{80}/.23/.\textbf{17} \\
\midrule
DiNO ViT-S/8 \cite{Sivanandan2023posh} & CP 1640 & .53/.12/.14 \\
\bottomrule
\end{tabular}
\label{table:poshcp-comparison}
\end{table}

\subsection{Predicting morphological features}\label{subsec:cp-pred}

\begin{figure}
    \centering
    \includegraphics[width=0.99\linewidth]{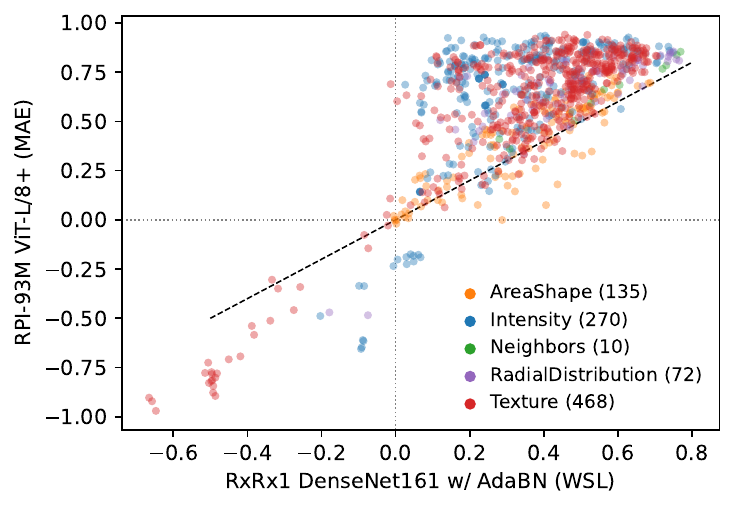}
    \caption{Single-task linear regression illustrates how an MAE-trained embedding model outperforms a WSL-trained model in predicting CellProfiler features across all categories.}
    \label{fig:cp_prediction}
\end{figure}

To determine whether models of different architectures were able to learn a diverse array of morphological characteristics, we used linear regression to predict 955 CellProfiler (CP) features spanning area-shape, texture, radial distribution, intensity, and neighbor categories \cite{10.1186/gb-2006-7-10-r100}. Although many of these features are highly correlated and display highly skewed distributions in practice, they nonetheless quantify a diverse set of specific morphological characteristics that can be used to assess the richness of model embeddings. Specifically, we observe that MAE model embeddings (RPI-93M ViT-L/8+) are better predictors of CP extracted morphological features than WSL model embeddings (RxRx1 DenseNet-161 w/ AdaBN), as measured by the coefficient of determination of predicted features from an independent experimental dataset (Fig. \ref{fig:cp_prediction}; see also Appendix~\ref{appendix:cellprofiler}). For example, improvements offered by this MAE over the WSL model range from a 14\% relative improvement in predicting the AreaShape features (.456 vs .401) to a 148\% improvement in predicting the Intensity feature (.737 vs .297), based on the median $R^2$. These observations suggest that MAEs can produce representations that more effectively capture a wide range of morphological features compared to the most performant WSL model proposed by \citet{sypetkowski2023rxrx1}.

\section{Conclusion}
\label{sec:conclusion}

This work demonstrates that scaling properties \cite{zhai2022scaling} apply to learning microscopy-based representations of cellular biology that can accurately infer known biological relationships. Unlike previous approaches that use weakly supervised learning \cite{Moshkov2022, sypetkowski2023rxrx1} on small, curated datasets, we show that the performance of self-supervised MAEs on biologically meaningful benchmarks scales to massive HCS image sets. Additionally, we introduce a novel reconstruction loss based on the Fourier transform which stabilizes large MAE training, and a channel-agnostic MAE architecture that generalizes to different channel configurations and offers promising directions for future work.

{
    \small
    \bibliographystyle{ieeenat_fullname}
    \bibliography{main}
}


\clearpage
\setcounter{page}{1}
\maketitlesupplementary
\appendix

\section{Appendix}

\subsection{Datasets} \label{appendix:datasets}
\textbf{RxRx1} \cite{sypetkowski2023rxrx1} is a publicly-available proprietary Cell Painting dataset with 125,510 images of 4 human cell types under 1,108 different siRNA perturbations across 51 experimental batches. 
A unique feature of this dataset is that it is comprised entirely of siRNA perturbations, which are known to have severe off-target effects silencing hundreds of genes~\cite{siRNAofftarget2010} causing very distinct phenotypes.

\textbf{RxRx1-2M} is a private version of RxRx1 containing over 1.6 million images across 16 different cell types and uses the same set of siRNA perturbations in RxRx1 from additional experimental batches.

\textbf{RxRx3} \cite{fay2023rxrx3} is a publicly-available proprietary Cell Painting dataset with over 2.2 million images of HUVEC cells each perturbed with one of 17,063 CRISPR knockouts (using one of six different guides) or 1,674 compounds across 180 experimental batches. This is the largest publicly available whole-genome HCS image set. 
CRISPR is a much more accurate technique for knocking out genes compare to siRNA and produces subtler phenotypes by targeting individual genes~\cite{barrangou2016applications}.

\textbf{RPI-52M} (Recursion Phenomics Imageset) is a private dataset with approximately 52 million proprietary images spanning 6,638 experimental batches and 40 cell types. This is a superset of the preceeding three datasets.

\textbf{RPI-93M} is a private dataset with approximately 93 million proprietary images spanning over 10,000 experimental batches and 41 cell types. To our knowledge, this is the largest HCS dataset collected for model training purposes. This is a superset of the preceding four datasets.

\textbf{Train and Validation splits}

All of the datasets are split such that model evaluation is performed on a non-overlapping set of \textit{experiments}, i.e. groups of multi-well plates containing replicates of perturbations in randomized layouts per plate, to avoid data-leakage.

\subsection{Model hyperparameters} \label{appendix:modelling}
Models trained on RxRx1 and RxRx1-2M were trained for 100 epochs, on RxRx3 for 50 epochs, and on RPI-52M and RPI-93M for up to 50 epochs, with early stopping depending on when validation performance plateaued. All models (except those using AdaBN) use random sampling without replacement over the full dataset to create training batches. Readers are encouraged to read \cite{sypetkowski2023rxrx1} for more details on batch construction for AdaBN models.

\subsubsection{Weakly supervised learning}
All WSL models were initialized from Image-Net pretraining weights. For the DenseNet-161-based classifiers, we searched over different batch sizes, learning rates, and optimizers. We empirically found that a batch size of 4,096 with standard SGD+momentum optimization performs best on the classification task, one-cycle learning rate schedule with cosine decay and a 10\% warm-up, a maximum learning rate of 0.32768, momentum of 0.9, and weight decay of 0.00001. For ViT-based classifiers, we used a batch size of 4,096, AdamW optimizer with a learning rate of at most 1e-3 using a one-cycle learning rate schedule with cosine decay and a 10\% warm-up, $\beta_1=$ 0.9 and $\beta_2=$ 0.95,  and a weight decay of 0.05. 

All non-AdaBN classifiers used weighted random sampling based on the perturbation labels in the dataset, whereas AdaBN models used a custom batch sampler to ensure that batches were sampled from the same experimental plate. For DenseNet-161-based classifiers, we used a sub-batch-size of 16 for GhostBN.

\subsubsection{Masked U-nets} \label{MUNets-appendix}
MU-Nets trained on RxRx3 used a global batch size of 4,096, while those trained on RPI-52M and RPI-93M used a global batch size of 16,384. Each was trained using the AdamW optimizer \cite{loshchilov2017decoupled} with $\beta_1=$ 0.9 and $\beta_2=$ 0.95, weight decay of 0.05, maximum learning rate 1e-3, cyclic cosine learning rate schedule, and no gradient clipping. We experimented with different mask ratios (25\%, 50\%, 75\%) and kernel sizes (3, 5). We compared the performance on the recall of biological relationships, similar to Table \ref{table:scaling}, for these values. Changing the mask ratio or kernel size did not seem to effect the performance. 

\subsubsection{Masked Autoencoder Vision Transformers}
MAE-ViTs on RxRx3 trained with a global batch size of 4,096, while those trained on RPI-52M and RPI-93M used a global batch size of 16,384. Each used the Lion optimizer~\cite{chen2023symbolic} with $\beta_1=$ 0.9 and $\beta_2=$ 0.95,  weight decay of 0.05, and no gradient clipping (based on the AdamW optimizer settings from \citet{he2022masked}). We found that training dynamics and downstream performance was significantly better with large batch sizes and the Lion optimizer versus using the recommended batch size and AdamW settings presented by \citet{SSLcookbook2023}. All ViT-S and \mbox{ViT-B} encoders were trained with a maximum learning rate of \mbox{1e-4} and all ViT-L encoders were trained with a maximum learning rate of 3e-5 (cosine decay schedule), based on initial experiments and recommended Lion learning rate settings presented in \cite{chen2023symbolic}. All MAE-ViTs were trained with stochastic depth \cite{SSLcookbook2023}, LayerScale \cite{SSLcookbook2023}, flash attention \cite{dao2022flashattention}, parallel scaling blocks \cite{dehghani2023scaling}, QK-normalization \cite{dehghani2023scaling}, and no QK-bias \cite{dehghani2023scaling}. Stochastic depth was set to 0.1 for ViT-S and ViT-B, and 0.3 for ViT-L. All models were initialized with random weights, as initial experiments found no benefit starting from pre-trained ImageNet weights.

\subsection{Training and Inference} \label{inference-appendix}

We scaled training based on the results of smaller models trained on smaller datasets \cite{dehghani2023scaling,hestness2017deep,openai2023gpt4,zhai2022scaling}, as visualized in Figure~\ref{fig:scaling} (total FLOps is based on \citet{touvron2022three}). Our most computationally intensive model, ViT-L/8+ (using the loss function described in Eq.~\ref{eq:combined}), was trained for over 20,000 GPU hours, learning on over 3.5 billion image crops sampled from RPI-93M. 

Models were trained with data-distributed parallel (DDP) training and PyTorch 2.0 for up to 100 epochs on up to 256 NVIDIA 80GB A100 GPUs, depending on the size of the model and dataset. 
256 x 256 x 6 image crops were randomly sampled from 2048 x 2048 x 6 images, augmenting with random horizontal and vertical flips. For each dataset, we use a validation set of center-cropped images from full experiments unseen during training. All image crops are preprocessed with channel-wise self-standardization \cite{sypetkowski2023rxrx1} before being passed into the deep learning models.

Inference was performed on a large-scale distributed kubernetes T4 GPU cluster. The results in Section~\ref{sec:results} are calculated on the gene knockout experiments of RxRx3~\cite{fay2023rxrx3}. 
Each \textit{well} in a biology experiment is loaded as a 2048 x 2048 x 6 int8 tensor. We tile over this image, obtaining 64 unique 256 x 256 x 6 crops. Each \textit{crop} is fed-forward through the encoder, and the resultant 64 embeddings are averaged to produce a final \textit{well-aggregated embedding}. Each genetics-only \textit{experiment} in RxRx3 has 9 plates, and each \textit{plate} has 1380 wells; therefore, nearly 800,000 samples need to be fed-forward through the encoder for each experiment. Given the 175 genetics-only experiments in RxRx3, this yields roughly 140 million individual samples fed-forward through each encoder in order to obtain genomic representations from the model. 
Note that the AdaBN-based weakly supervised models require careful mini-batch construction during both training and inference, whereas the rest of our models are deterministic in producing embeddings of individual samples.

\subsection{Additional reconstructions} \label{appendix:recons}
\begin{figure*}[ht]
    \centering
    \begin{minipage}{.3333\textwidth}
        \centering
        \includegraphics[width=\linewidth]{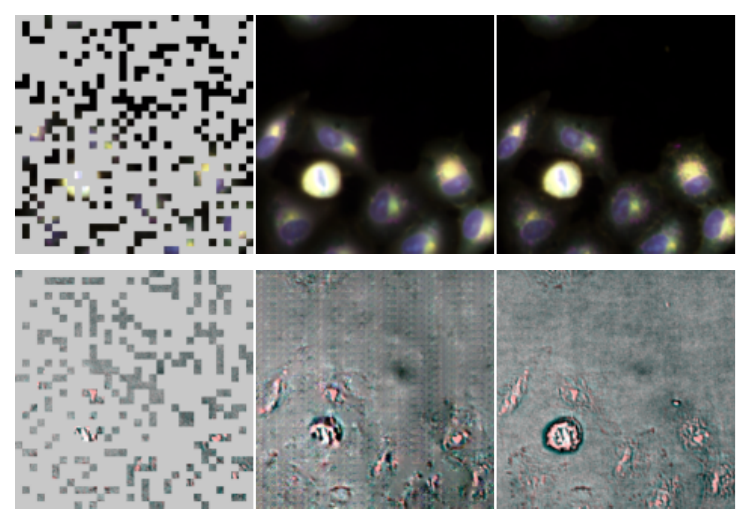}
    \end{minipage}%
    \begin{minipage}{.3333\textwidth}
        \centering
        \includegraphics[width=\linewidth]{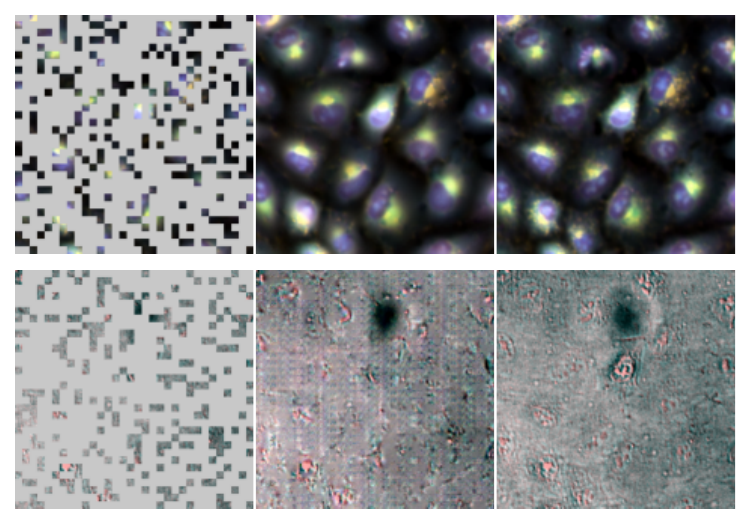}
    \end{minipage}%
    \begin{minipage}{.3333\textwidth}
        \centering
        \includegraphics[width=\linewidth]{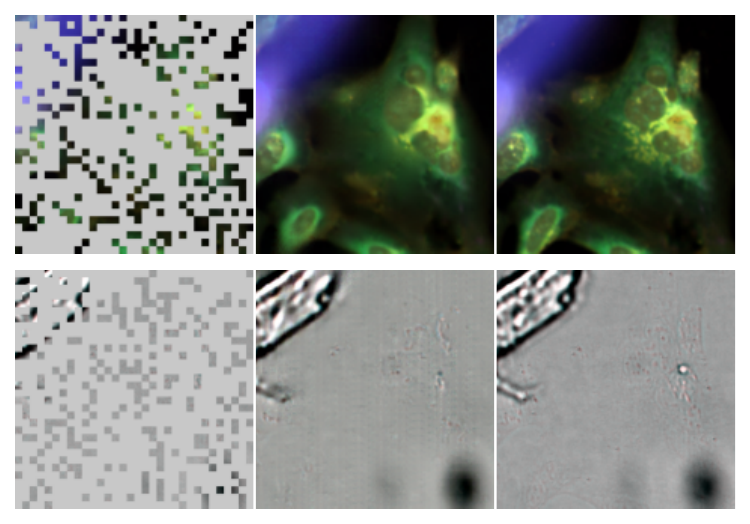}
    \end{minipage}
    \caption{Visualizing MAE ViT-L/8+ (trained on RPI-93M) 75\% masked reconstructions on randomly selected out-of-domain JUMP-CP \cite{chandrasekaran2023jump} image crops. Rows alternate between Cell Painting and Brightfield images obtained from the same well. Note that the wells in JUMP-CP were imaged using different assays, channel composition, microscopes, and labs compared to the well images we used for pre-training.}
    \label{fig:jump_recon}
\end{figure*}

Additional visualizations of the reconstructed masked input images using MAE ViT-L/8+ on the JUMP-CP dataset, for both Cell Painting and Brightfield channels, are shown in Figure~\ref{fig:jump_recon}. Recall that JUMP-CP was not included in any training set, thus this data is OOD. Nevertheless, the MAE reconstruction generalizes well to this dataset, especially for the Cell Painting samples.

\subsection{Additional results} \label{sec:additional}

\begin{figure*}
    \centering
    \begin{minipage}{0.49\textwidth}
        \centering
        \includegraphics[width=0.99\textwidth]{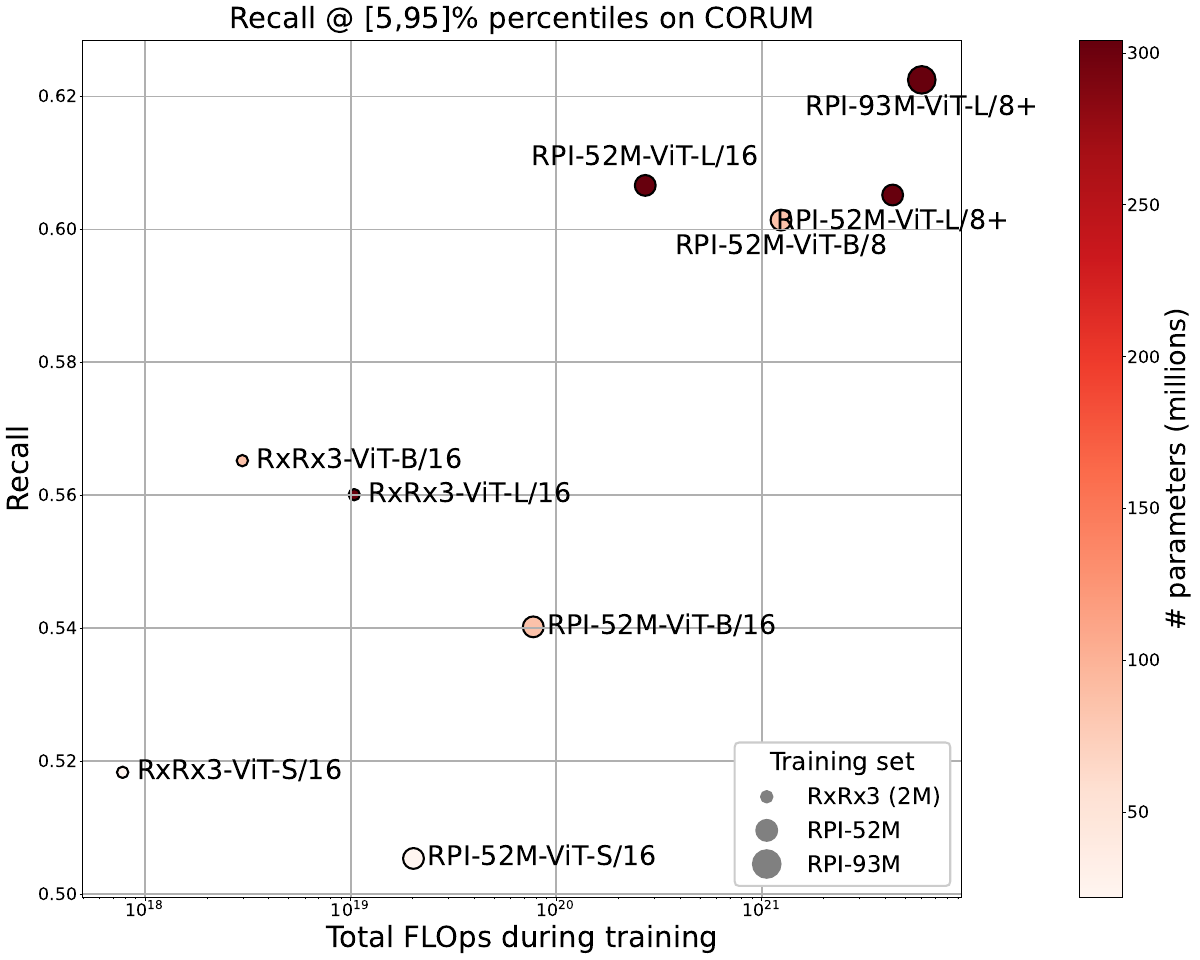}  
    \end{minipage}%
    \begin{minipage}{0.02\textwidth}
        \ 
    \end{minipage}%
    \begin{minipage}{0.49\textwidth}
        \centering
        \includegraphics[width=0.99\textwidth]{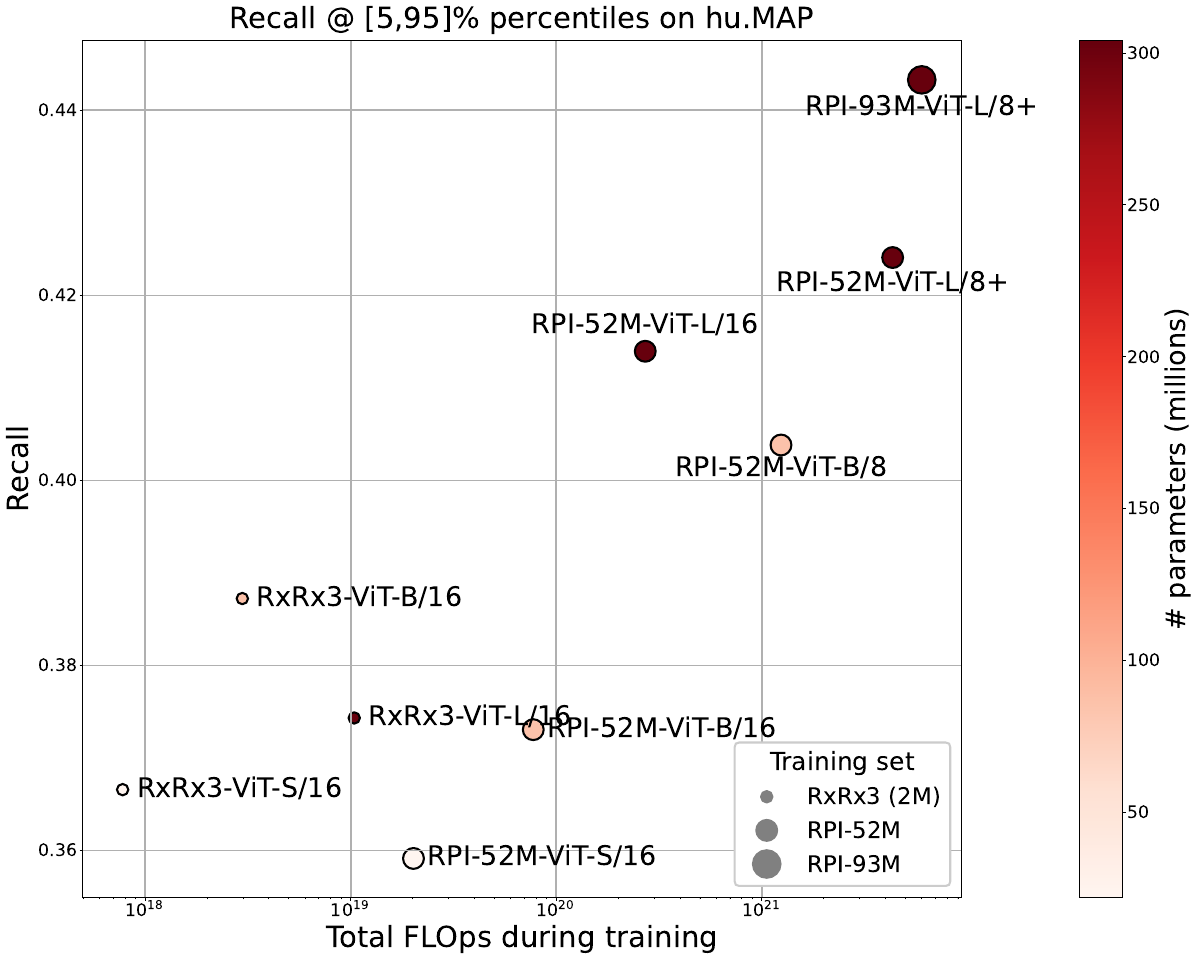} 
    \end{minipage}
    \centering
    \begin{minipage}{0.49\textwidth}
        \centering
        \includegraphics[width=0.99\textwidth]{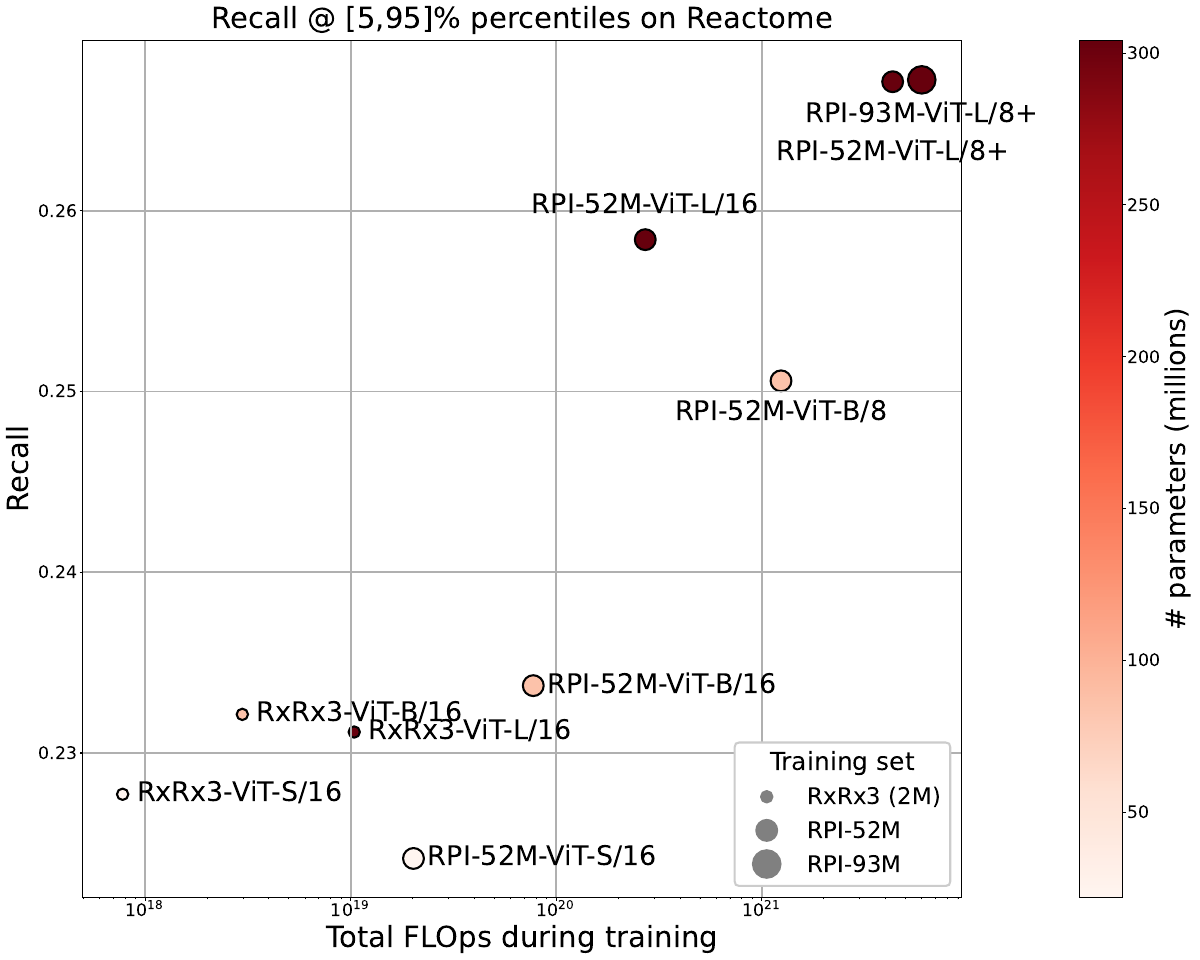}  
    \end{minipage}%
    \caption{CORUM, hu.MAP, and Reactome recalls for ViTs as a function of training FLOps.}
    \label{fig:scaling_others}
\end{figure*}

\textbf{Calculation of FLOps}. In Figure~\ref{fig:scaling_others} we include the scaling plots as in Figure~\ref{fig:scaling}, for the other three benchmark databases (CORUM, hu.MAP, and Reactome). Floating point operations (FLOps) are approximated based on the FLOp counts presented in Table 1 from \citet{touvron2022three}, which presents FLOps for ViT-S/B/L/16 on a 224x224x3 image. We adjust flop counts by a factor of $(\frac{16*16}{14*14})^2 = 1.69$ to account for the changed crop size, and then for 8x8 patching models we multiply by a factor of 16 to account for the 4x more tokens and the quadratic impact this has on the attention head computations. We lastly multiply the FLOps by the number of image crops seen during training for each model.

\subsection{CellProfiler feature prediction} \label{appendix:cellprofiler}
\begin{figure*}
    \centering
    \includegraphics[width=0.99\linewidth]{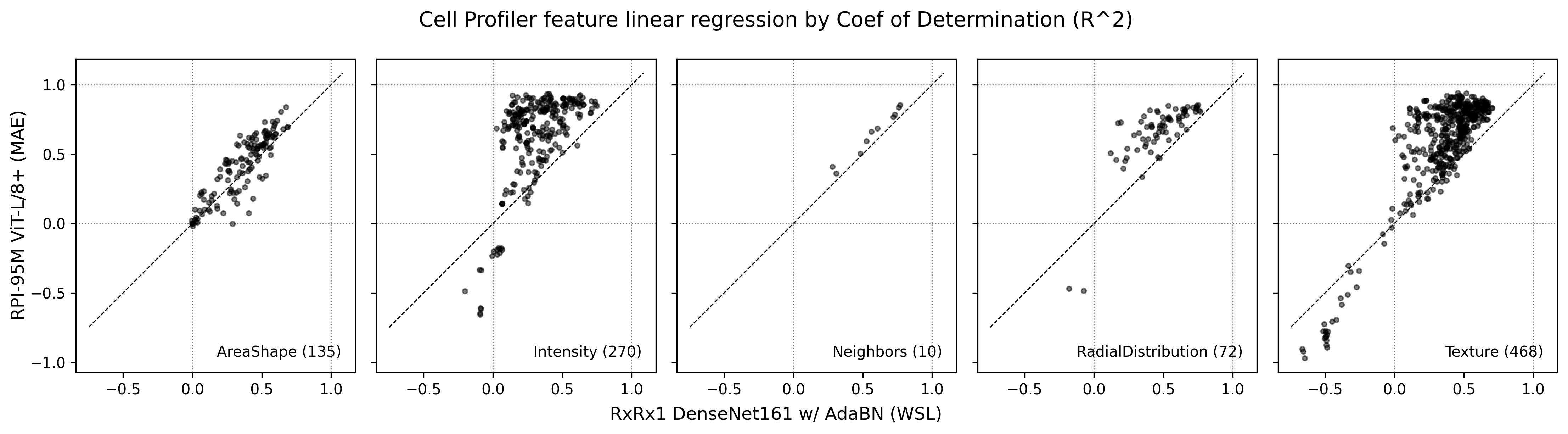}
    \caption{Single-task linear regression illustrates how an MAE-trained embedding model outperforms a WSL-trained model in predicting CellProfiler features across all categories.}
    \label{fig:cp_prediction_expand}
\end{figure*}

We tested the ability of two models and model architectures, RxRx1 DenseNet-161 w/ AdaBN (WSL), and RPI-93M ViT-L/8+ (MAE) to predict CellProfiler (CP) features using linear regression. Training was performed on one internal experiment representing 12 plates of 1380 wells each, for a total of 16,560 wells. Testing was performed with a different internal experiment of the same size representing 1,160 different CRISPR knock-out perturbations (with 121 control perturbations in common, equaling ${<}$ 10\% reagent overlap between train and test experiments). 955 CP features were extracted over the categories of area-shape, intensity, neighbors, radial-distribution, and texture, and averaged to the well-level. Highly-skewed CP feature distributions were transformed by log scaling (skew ${>}$ 0.5) or by squaring (skew ${<}$ -0.5) to make them more normal then all features were centered to 0 and scaled to unit variance. 1,024-dimensional embeddings for both models were similarly averaged to the well-level, centered to 0, and scaled to unit variance. All feature predictors were trained as single-task linear regressors using scikit-learn’s ElasticNetCV estimator class. A grid-search over a small range of L1/L2 ratios (0.1, 0.6, 0.9, 0.95, 0.99) and alphas (auto-determined) with a 5-fold cross-validation schedule was used. The best-fit parameters were then used to predict and score the independent experiment test set using the coefficient of determination (Fig. \ref{fig:cp_prediction}, Supp. Fig. \ref{fig:cp_prediction_expand}, Supp. Table \ref{tab:supp-cp-prediction}).

\subsection{JUMP-CP benchmarks} \label{appendix:jumpcp-results}
The un-aggregated data for Table~\ref{table:jumpcp-benchmark} are presented in Table~\ref{tab:jumpcp-detection} and Table~\ref{tab:jumpcp-sibling}.

\begin{table*}[t]
\centering
\caption{Summary of results discussed in Section~\ref{sec:results}, including additional results for smaller models. Recall of known relationships in top and bottom 5\% of cosine similarities by model, training set, and database (CORUM/hu.MAP/Reactome/StringDB).} 
\begin{tabular}{lcccc}
\toprule
\textbf{Model backbone} / Pretraining dataset & RxRx1 \cite{sypetkowski2023rxrx1} & RxRx3 \cite{fay2023rxrx3} & RPI-52M & RPI-93M \\
\midrule
\textit{WSL}\\
DenseNet-161 & .383/.307/.190/.330 & .359/.271/.174/.319 & -- & -- \\
DenseNet-161 w/ AdaBN & .485/.349/.228/.417 & .461/.303/.188/.377 & -- & -- \\
DenseNet-161 w/ AdaBN (1024-dim) & .502/.363/.220/.422 & .520/.350/.207/.413 & -- & -- \\
\midrule
\textit{SSL models} \\
MU-net-M  & -- & .557/.382/.236/.432 & -- & -- \\
MU-net-L  & -- &.566/.374/.232/.427 & .576/.385/.238/.443 & .581/.386/.247/.440 \\
MAE ViT-S/16 & -- & .518/.367/.228/.415 & .505/.359/.224/.402 & -- \\
MAE ViT-B/16 & -- & .565/.387/.232/.435 & 540/.373/.234/.416    & -- \\
MAE ViT-B/8  & -- & -- & .601/.404/.251/.459 & -- \\
MAE ViT-L/16 & -- & .560/.374/.231/.427 & .607/.414/.258/.460 & -- \\
MAE ViT-L/8+ & -- & -- & .605/.424/\textbf{.267}/.474 & \textbf{.622}/\textbf{.443}/\textbf{.267}/\textbf{.484} \\
\end{tabular}
\label{table:scaling}
\end{table*}

\begin{table*}[ht]
\centering
\caption{Median $R^{2}$ ($\pm$ median absolute deviation) for CellProfiler predictions across feature categories.}
\label{tab:supp-cp-prediction}
\begin{tabular}{lccccc}
\toprule
\textbf{Model Backbone} & AreaShape & Intensity & Neighbors & RadialDistribution & Texture \\
\midrule
RxRx1 DN161 w/ AdaBN (WSL) & 0.401 \small$\pm$0.127	& 0.297 \small$\pm$0.121	& 0.583 \small$\pm$0.142	& 0.484 \small$\pm$0.127	& 0.413 \small$\pm$0.112 \\
RPI-93M ViT-L/8+ (MAE) & 0.456 \small$\pm$0.162	& 0.737 \small$\pm$0.120	& 0.674 \small$\pm$0.137	& 0.711 \small$\pm$0.093	& 0.705 \small$\pm$0.133 \\
\bottomrule
\end{tabular}
\end{table*}

\begin{table*}[ht]
    \centering
    \caption{Perturbation retrieval on the JUMP-CP dataset, measured in fraction retrieved.}
    \label{tab:jumpcp-detection}
    \begin{tabular}{cccccccc}
    \multicolumn{3}{c}{}&\multicolumn{4}{c}{\textbf{Model backbone, dataset}} \\ 
    \cmidrule{4-7}
    
         Cell type&Modality&Time-point& CA-93M-ViT-L & CA-93M-ViT-L-8chans & ViTL-Image-net & cellprofiler \\ 
        \midrule
        \multirow{6}{*}{A549}&\multirow{2}{*}{compound}&long & \textbf{1.00} & 0.99 & 0.99 & 0.95 \\ 
        &&short & 0.98 &\textbf{ 0.99} & 0.93 & 0.76 \\ 
        &\multirow{2}{*}{crispr}&long & 0.89 & \textbf{0.95} & 0.90 & 0.68 \\ 
        &&short & 0.88 &\textbf{ 0.97} & 0.90 & 0.68 \\ 
        &\multirow{2}{*}{orf}&long & \textbf{0.84} & 0.83 & 0.71 & 0.05 \\ 
        &&short & 0.63 & \textbf{0.93} & 0.78 & 0.06 \\ 
        \midrule
        \multirow{6}{*}{U2OS}&\multirow{2}{*}{compound}&long & 0.98 & \textbf{0.99} & 0.94 & 0.66 \\ 
        &&short & 0.88 & \textbf{0.97} & 0.88 & 0.78 \\ 
        &\multirow{2}{*}{crispr}&long & 0.91 & \textbf{0.96} & 0.94 & 0.46 \\ 
        &&short & 0.91 & \textbf{0.98} & 0.94 & 0.67 \\ 
        &\multirow{2}{*}{orf}&long & 0.65 & \textbf{0.89} & 0.75 & 0.20 \\ 
        &&short & 0.79 & 0.89 & \textbf{0.90} & 0.37 \\ 
        \bottomrule
    \end{tabular}
\end{table*}

\begin{table*}[ht]
    \centering
    \caption{Siblings retrieval on the JUMP-CP dataset, measured in fraction retrieved. Note that ORF's do not have siblings.}
    \label{tab:jumpcp-sibling}
    \begin{tabular}{cccccccc}
    \multicolumn{3}{c}{}&\multicolumn{4}{c}{\textbf{Model backbone, dataset}} \\ 
    \cmidrule{4-7}
    
         Cell type&Modality&Time-point& CA-93M-ViT-L & CA-93M-ViT-L-8chans & ViTL-Image-net & cellprofiler \\ 
        \midrule
        \multirow{4}{*}{A549}&\multirow{2}{*}{compound}&long & 0.05 & 0.04 & 0.13 & \textbf{0.17} \\ 
        &&short & 0.13 & 0.04 & 0.08 & \textbf{0.14} \\ 
        &\multirow{2}{*}{crispr}&long & 0.06 & 0.01 & 0.07 & \textbf{0.12} \\ 
        &&short & 0.04 & 0.01 & 0.04 & \textbf{0.11} \\ 
        \midrule
        \multirow{4}{*}{U2OS}&\multirow{2}{*}{compound}&long & 0.12 & 0.00 & 0.03 &\textbf{ 0.25} \\ 
        &&short & 0.06 & 0.02 & \textbf{0.05} & 0.04 \\ 
        &\multirow{2}{*}{crispr}&long &  0.03 & 0.02 & 0.03 & \textbf{0.18}  \\ 
        &&short & 0.03 & 0.02 & 0.02 & \textbf{0.07} \\ 
        \bottomrule
    \end{tabular}
\end{table*}

\end{document}